\documentclass[doc,floatsintext]{apa6}

\usepackage[english]{babel}

\usepackage{amssymb}
\usepackage{siunitx}
\PassOptionsToPackage{hyphens}{url}\usepackage{hyperref}
\usepackage{cleveref}
\usepackage[utf8]{inputenc}
\usepackage[right]{lineno}
\usepackage{csquotes}
\usepackage{booktabs}
\usepackage{longtable}
\usepackage{adjustbox}
\usepackage{array}
\usepackage{url}
\usepackage{authblk}
\usepackage{xcolor} 

\usepackage{natbib}

\usepackage[english]{babel}

\crefformat{figure}{#2Figure~#1#3}
\Crefformat{figure}{#2Figure~#1#3}
\crefformat{table}{#2Table~#1#3}
\Crefformat{table}{#2Table~#1#3}
\crefformat{section}{#2Section~#1#3}
\Crefformat{section}{#2Section~#1#3}

\def\fp{F\&P}
\def\fps{F\&P\ }

\author{
Thomas L.~Griffiths$^{1*}$, Brenden M.~Lake$^{1*}$, R.~Thomas McCoy$^{2*}$, \\ \vspace{-4mm} Ellie Pavlick$^{3*}$, Taylor W.~Webb$^{4*}$
}

\affiliation{
$^{1}$\mbox{Departments of Psychology and Computer Science, Princeton University},
$^{2}$\mbox{Department of Linguistics and Wu Tsai Institute, Yale University},
$^{3}$\mbox{Department of Computer Science, Brown University},
$^{4}$\mbox{Department of Psychology, Université de Montr\'{e}al and Mila -- Quebec AI Institute}, $^{*}$\mbox{All authors contributed equally, order is alphabetical}
}

\title{{\bf Whither symbols in the era of \\ advanced neural networks?}}

\shorttitle{Whither symbols?}

\abstract{%
Some of the strongest evidence that human minds should be thought about in terms of symbolic systems has been the way they combine ideas, produce novelty, and learn quickly. We argue that modern neural networks -- and the artificial intelligence systems built upon them -- exhibit similar abilities. This undermines the argument that the cognitive processes and representations used by human minds are symbolic, although the fact that these neural networks are typically trained on data generated by symbolic systems illustrates that such systems play an important role in characterizing the abstract problems that human minds have to solve. This argument leads us to offer a new agenda for research on the symbolic basis of human thought.
}

\begin{document}
\maketitle



\begin{figure}[t!]
    \centering
    \includegraphics[width=1.0\linewidth]{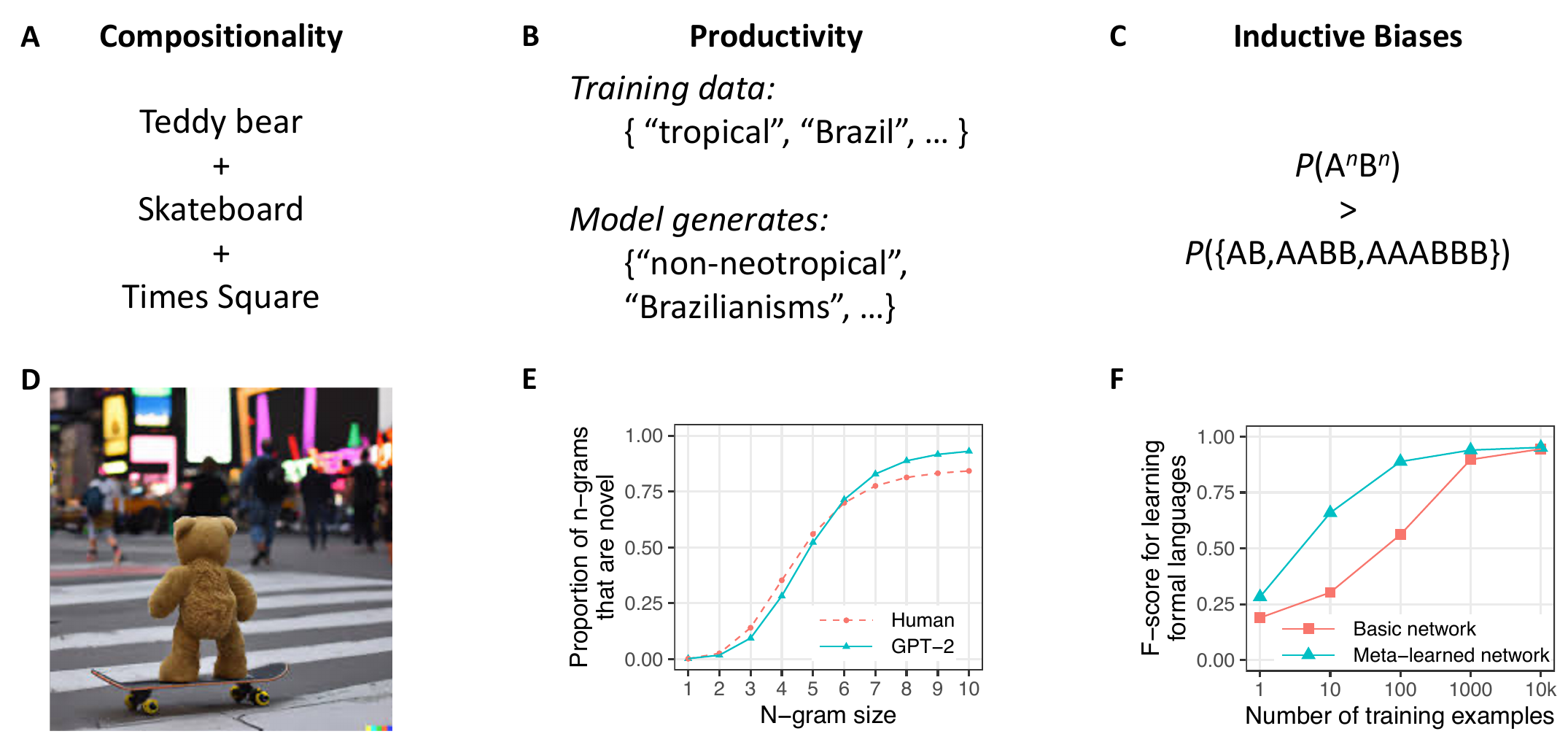}
    \caption{Signatures of symbol processing in modern neural networks. The figure illustrates three properties that have traditionally been taken as evidence for symbol processing in the human mind. \textbf{A.} Compositionality refers to the ability to flexibly construct thoughts out of a set of simpler elements. \textbf{B.} Productivity refers to the ability to extend observed processes to new examples, generating a potentially unboundedly large number of novel thoughts or sentences.  \textbf{C.} Inductive biases refer to the set of assumptions that guide how a learner learns from training data. \textbf{D.} An image of ‘a teddy bear on a skateboard in times square’, generated by the text-to-image model DALL-E 3. Similar examples are ubiquitous in the outputs of modern generative AI systems, illustrating a capacity for compositionality. \textbf{E.} Evidence for productivity in large language models. The language model GPT-2 produces novel n-grams at a rate similar to the rate seen in a baseline of human-generated text \cite{mccoy2023raven}. \textbf{F.} Evidence for symbolic inductive biases in neural networks: with a training approach based on meta-learning, a neural network can learn symbolic formal languages from small numbers of examples \cite{mccoy2025modeling}. 
}\label{fig:main_figure}
\end{figure}

\noindent
Questions about whether artificial neural networks can represent symbolic systems such as logic and grammar have been discussed for almost as long as neural networks have been used to model human cognition \cite{mcculloch1943logical,minsky1961steps,Rumelhart1986past,Pinker1988language}. Properties of symbolic systems --- productivity and compositionality (Figure~\ref{fig:main_figure}) --- have been proposed as central to human intelligence \cite{chomsky1957syntactic,Fodor1988,Quilty-Dunn2023}, so neural networks' alleged inability to demonstrate these properties is potentially a serious critique. Moreover, probabilistic models that operate over logic, grammars, or programs have highlighted how symbolic systems can be used to characterize the inductive biases that allow people to learn from limited data \cite{Griffiths2010tics,griffiths2024bayesian}.

In this article, we revisit these questions in light of recent advances in AI:   training larger neural networks with more data   using new architectures and training techniques has resulted in improvements in
reasoning, learning, and using language \cite{kaplan2020scaling}. For instance, large language models (LLMs) \cite[e.g.,][]{GPT4techreport,grattafiori2024llama,groeneveld2024olmo,anthropic2024claude} based on the transformer architecture \cite{Vaswani2017attention} display impressive  productivity, and the training technique of meta-learning \cite{schmidhuber1987evolutionary,thrun2012learning} results in greater compositionality and rapid learning \cite{LakeMLC,mccoy2025modeling}. 

We argue that these advances undermine historical arguments that human minds use symbolic systems in ways that neural networks do not. Neural networks display comparable compositionality, productivity, and inductive biases to humans when evaluated against the same empirical standards as humans. Therefore, further progress in understanding the similarities and differences between human cognition and neural networks will require new methods for evaluating such claims. 

These recent advances  also suggest a way to resolve this classic debate, via levels of analysis (see Box 1). These demonstrations of compositionality, productivity, and rapid learning arise in neural networks that have been trained on data generated by symbolic systems.
 Similarly, symbolic systems also feature strongly in human experience (both within and across lifespans). Thus, for both humans and machines, symbolic systems are useful in characterizing the abstract computational problems that need to be solved --- what Marr \cite{Marr1982} called the computational level. The subsymbolic processes and representations that implement these computations in neural networks play a complementary role at Marr's algorithmic level and capture other behaviors that defy easy symbolic description.

In the remainder of this article, we review innovations behind modern neural networks and discuss three classic domains of evidence for symbolic systems: compositionality, productivity, and inductive biases. In each domain we summarize results demonstrating the capacities of neural networks. We argue that while symbolic systems remain  valuable in characterizing and interpreting human cognition at the computational level, neural networks  provide an increasingly complete account of how human minds and brains approximate those symbolic systems. Evaluating this hypothesis will require a new research agenda with four interrelated threads: designing more diagnostic tasks for probing symbolic representations, developing a deeper mechanistic understanding of neural networks, training neural networks in more developmentally plausible ways, and creating more explicitly ``cognitive'' neural networks that explain human mental processes. 

\section{Advances behind modern neural network systems} \label{sec_advances}

The basic idea behind artificial neural networks has remained the same since their origins~\cite{mcculloch1943logical,Rosenblatt1958}: Units simulating neurons are connected  by weighted links. The links are directed, and each unit's activation depends on the sum of the weighted activations of the units that feed into it. The weights are adjusted via a learning algorithm (most notably backpropagation~\cite{Rumelhart1986}), steering the system towards producing appropriate outputs for particular inputs. 

Several factors underlie the successes of modern neural networks. The most obvious  is their scale. Modern neural networks can have millions, billions, or trillions of parameters representing weighted connections. Size is an important factor in allowing these systems to perform a wide range of tasks~\cite{kaplan2020scaling}. Increases in  size have been accompanied by increases in the amount of training data~\cite{hoffmann2022training}. Modern large language models are routinely trained on trillions of linguistic tokens (words, parts of words, or symbols). The size and breadth of these data support the increasingly general capabilities of the resulting models.

These increases in scale have been accompanied by advances in training procedures. Large language models are trained to predict the next token given a sequence of preceding tokens. This leads to a capacity for in-context learning: the ability to learn new tasks on the fly based on a small set of examples presented as text~\cite{Brown2020}. This can be viewed as a form of meta-learning (see Box 2).  Other forms of meta-learning can be used to train neural networks with human-like capacities for compositional generalization~\cite{LakeMLC} and rapid learning \cite{mccoy2025modeling}. New training techniques also include methods for enhancing instruction-following~\cite{bai2022training} and optimization to perform explicit sequential reasoning using text (referred to as a ``chain of thought'').

New architectures, such as the transformer \cite{Vaswani2017attention}, were a major driver of LLMs and of subsequent progress in vision~\cite{Dosovitskiy2020image}. Compared to the recurrent neural networks~\cite{Hochreiter1997} previously used in language modeling, transformers can better learn the long-range dependencies that are critical for understanding language. Transformers and other external memory architectures~\cite{Graves2014,Graves2016,Webb2021emergent} can be viewed as implementing the memory-addressing procedures that are central to symbolic computing: information at one position in a sequence can direct attention to another position in the sequence. This enables transformers to dynamically recombine knowledge, a key property of symbolic systems. Consistent with this, transformers seem particularly well suited to in-context learning~\cite{Chan2022data,Akyurek2024icl}, and have played an important role in meta-learning approaches for compositional generalization~\cite{LakeMLC}.

Together, these advances in training procedures and architectures, along with dramatic increases in model size and training data, have produced systems that can fluently generate natural language~\cite{Piantadosi2023refute} and that are increasingly capable of human-like reasoning and generalization~\cite{webb2023emergent,chollet2024o3,Marinescu2024distilling}, capacities that have previously been thought to require mechanisms for explicit symbol processing. The following sections  discuss these capabilities, illustrating how they satisfy criteria offered as arguments in favor of symbolic systems.

\section{Compositionality}
An influential argument for symbolic systems focuses on systematic compositionality -- the algebraic capacity to understand or produce novel combinations of known components. Systematically compositional systems show symmetries in thought and behavior: entertaining one thought implies the ability to entertain related thoughts \cite{Fodor1988}. For instance, understanding the sentence ``dogs chase cats'' implies one can understand ``cats chase dogs''; it would be odd to find a human with  ``piecemeal cognition'' for which this didn't hold. 

Fodor and Pylyshyn \cite{Fodor1988} (hereafter \fp) famously argued that because traditional neural networks don't have systematic compositionality, they aren't viable models of the mind. In contrast, symbolic models assume this property by construction.  \fp's article led to a long debate  \cite{smolensky1991constituent,Marcus2003,KentJohnson2004,Symons2014,Hill2020environmental,Quilty-Dunn2023}, with two common reactions from critics: specialized neural networks can be more systematic than \fps claimed \cite{Smolensky1990,Pollack1990,Kriete2013,Chalmers1990why}, and  humans can be less systematic than \fps claimed  \cite{KentJohnson2004,OReilly2014}. More recent work has revisited \fp's arguments in light of modern neural networks, showing that new architectures alone don't address the issue; that is, training on perfectly systematic data doesn't lead to systematic generalizations~\cite{LakeBaroni2018generalization,Ettinger2018assessing,Kim2020cogs,Hupkes2020}.

The evidence for compositionality is more positive in LLMs. Anecdotally, interacting with LLMs doesn't bring to mind the ``piecemeal cognition'' that worried \fp. Today's LLMs can handle a great variety of tasks and topics, and do not exhibit extreme sensitivity to wording or formatting. LLMs can compose concepts  that are unlikely to have appeared together during training \cite{Yu2023skillmix}, e.g., writing a Shakespearean poem proving there are infinitely many primes  \cite{Bubeck2023sparks}. LLMs can also produce sentences that are genuinely novel in terms of word sequence or syntactic structure \cite{mccoy2023raven}. Finally, LLMs are increasingly fluent in  programming languages that are, of course, paradigmatic symbolic systems. 

There is also evidence for compositionality in vision-language models (VLMs) that handle visual inputs and outputs. Anecdotally, VLMs can generate unusual combinations, e.g., a person composed of the letters O, H, and Y \cite{Bubeck2023sparks} or a teddy bear riding a skateboard in Times Square \cite{Ramesh2022hierarchical} (Figure \ref{fig:main_figure}D). These abilities are imperfect; failures can relate to binding objects and properties \cite{Ramesh2022hierarchical,CampbellBinding,lewis2024does}, combining many attributes together \cite{ConceptMix2025conceptmix,FuLeeComp2025evaluating}, or handling statistically rare combinations \cite{LoveringPavlick2023}. It is difficult to compare the output of humans and machines directly, but on the task of verifying whether a generated image follows its compositional specifications, people and VLMs have been found to be comparable \cite{ConceptMix2025conceptmix}.

Where did this compositionality come from? Scale is certainly a factor, but there is also evidence neural networks show enhanced capabilities with the right incentives and practice \cite{IriePIP}. Specifically, meta-learning can be used to enhance compositionality: it can help a network handle familiar inputs in novel, compositional contexts \cite{LakeMLC} (see Box 2; also see \cite{Chan2022data} for the connection between scale and meta-learning).

This method has led to models (even at a small scale) that show more human-like capacities for inferring and composing rules \cite{LakeMLC,russin2024human}, for learning categories based on compositional building blocks \cite{Marinescu2024distilling}, and for learning to use new words from just a few examples \cite{WangMinnow}. Importantly, mimicking human behavior in these settings often diverges from maximizing systematicity. Empirical studies suggest that people use heuristics  that deviate from symbolic ideals, such that meta-learning neural networks (rather than purely symbolic models) best capture the heterogeneity of human behavioral responses \cite{LakeMLC,ZhouFeinman2023}.

Despite increasing compositionality in neural networks,  challenges remain. Small changes to the input can still change the behavior of networks in unexpected ways. For instance,  capitalization, misspellings, and tone changes can lead LLMs to respond unsafely to dangerous inputs \cite{HughesBestoN,SageEval}, behavior that is more piecemeal than systematic. LLMs are also influenced by the statistics of their training data in ways that can interfere with more systematic forms of reasoning \cite{mccoy2024embers,Berglund2023reversal}. 
The human-machine gap is less clear here: people are also sensitive to priors and can be intuitive rather than algebraic reasoners. For example,  both humans and LLMs allow world knowledge to hinder judgments of logical validity \cite{Dasgupta2022}. Overall, we still see a compositionality gap between neural network and human behavior, but the gap is smaller than ever before. Moreover, because humans are also imperfect at rule-like, compositional generalization, neural networks that are trained to engage in symbolic behavior \cite{SymbolicBehavior} are likely to provide a more precise algorithmic-level account (Box 1) than more purely symbolic alternatives.

\section{Productivity}

A property that is closely related to compositionality is productivity -- the capacity to produce or process structures that have never been encountered before. This property is a consequence of compositionality,
which naturally supports novel arrangements of symbols.
Productivity is a central property of language \cite{hockett1960origin}: people routinely say completely new sentences.
The productivity of language is typically taken to be unbounded, with the capacity to make ``infinite use of finite means'' \cite{humboldt1836uber,chomsky1965aspects} by creating and comprehending an unlimited number of possible sentences built from a finite vocabulary: \textit{It's raining}, \textit{She knows that it's raining}, \textit{He believes that she knows that it's raining}, etc. 
\fps identified productivity as a characteristic property of cognitive architectures that are based on symbolic systems \cite{Fodor1988}.

Recent LLMs possess productive abilities. 
First, as discussed in the previous section, they routinely display the most basic type of productivity -- the ability to produce and process novel structure -- since they frequently produce sentences that were not in their training data. 
This novelty is at the level of both word sequences and syntax; for example, about 67\% of GPT-2-generated sentences have syntactic structures that no sentence in the training set shares \cite{mccoy2023raven}.
The production of novel structures could potentially be explained away as resulting from errors (e.g., \textit{the dog are barking} would likely be novel), but LLMs generally produce \cite{mccoy2023raven} and process \cite{gulordava2018colorless,Wei2021} novel text in ways that satisfy the rules of syntax, albeit with factual errors (``hallucinations'':  \cite{wiseman2017challenges,ji2023survey,huang2025survey}).
Thus, LLMs 
capture a meaningful degree of productivity in at least one area (syntax), even if hallucinations show that these abilities do not extend as robustly to semantics.

On a deeper level, productivity is often associated not just with novelty but with \textit{unbounded} novelty---the ability to handle unboundedly many structures. 
Do neural networks possess this deeper type of productivity? Many standard neural networks show limitations in generalizing to long sequences \citep{LakeBaroni2018generalization}.
However, humans also struggle on some long inputs, such as sentences featuring deep center embedding, a fact that has been argued to reflect memory limitations rather than constraints on productivity \cite{miller1963finitary}; 
similar arguments may apply to at least some generalization failures in neural networks \cite{lampinen2024language,gong2024working,frankland2021no}.
Further, techniques have been developed that substantially improve length generalization in neural networks, such as specialized positional embeddings \cite{kazemnejad2023impact,wang2024length} and meta-learning from an appropriate distribution of formal languages
\cite{mccoy2025modeling} (see Box 2).
Thus, neural networks can perform fairly robust length generalization in at least some situations,  
providing an existence proof that neural networks can display an extensive degree of productivity. 

\section{Inductive biases}

One of the most salient lines of criticism of LLMs is their data inefficiency, with their linguistic abilities arising from trillions of words of pretraining data. In contrast, human learners are estimated to require orders of magnitude less data, on the order of 10 million to 100 million words \cite{Linzen2020accelerate,Warstadt2022,frank2023bridging}. When LLMs are instead trained on such data, their linguistic abilities are no longer as impressive \cite{vanschijndel2019quantity,Zhang2021c,yedetore2023poor}. 

Data inefficiency ultimately reflects inductive biases. That is, the fact that LLMs need so much data is evidence that they lack human-like inductive biases -- those factors that lead a learner to favor some solutions over others. The data presented to any learner (human or AI) will be consistent with many possible solutions, and thus inductive biases are needed to select a solution.  When told that the string \texttt{1,2 -> 3} is valid and asked what other strings might be valid, one learner might guess \texttt{3,4 -> 5},  reflecting a bias toward reasoning over the number line. Another learner might guess \texttt{3,4 -> 3}, reflecting a bias that strings with characters matching the original example are more likely to be valid. The way that a computational model generalizes is thus a way to evaluate its cognitive plausibility.

Humans are known to enter the world with powerful inductive biases. For example,  babies seem to have biases toward recognizing and tracking objects \cite{Spelke2007} and toward inferring causal relationships in the world \cite{Gopnik2004theory}. Until recently, only cognitive models that use symbolic representations could reproduce human-like inductive biases. For example, work on the probabilistic language of thought \cite{Fodor1975,Goodman2008a,Piantadosi2016c} and on probabilistic program induction \cite{Rule2018learning,Rule2020} has argued that strong biases toward symbolic primitives based on logic, geometry, or physics are necessary to match human capacities. Neural network models trained from scratch  often demonstrate a lack of compositionality \cite{LakeBaroni2018generalization,Kim2020cogs,pavlick2022semantic} or use of ``shallow'' heuristics \cite{mccoy2019right,poliak2018hypothesis}, motivating explanations of human inferences based on ``language of thought'' style hypotheses even in domains such as vision \cite{Sable-Meyer2021,Quilty-Dunn2023} that are typically assumed to be more amenable to neural network computation than domains such as language and logic. In language, inductive biases for hierarchy and recursion \cite{chomsky1995minimalist,Hauser2002}  have been seen as necessary to overcome the otherwise weak inductive biases of neural networks  \cite{Linzen2016assessing,gulordava2018colorless}, motivating the creation of neural networks that incorporate such biases \cite{Bowman2015Tree,socher2013recursive}. 

Recent advances in LLMs challenge this view. LLMs capable of in-context learning   appear to exhibit rapid few-shot learning \cite{Brown2020}, human-level analogical reasoning \cite{webb2023emergent}, logical concept learning \cite{loo2025llms}, and compositional generalization \cite{russin2024human}. As was the case with compositionality and productivity, these apparent discrepancies between standard neural network learning and LLMs are often associated with meta-learning. For example, neural networks can be meta-trained to encode Bayesian priors \cite{mccoy2025modeling,Marinescu2024distilling} and mirror human-like curriculum effects and error patterns \cite{russin2024curriculum}. Results of this nature have been used to argue that pretrained neural networks must be treated as an entirely different beast -- in terms of their architecture and learning biases -- than their trained-from-scratch counterparts \cite{lovering2021predicting,campbell2024human}

Taken together, this recent evidence points to an understanding of inductive biases in neural networks that is more nuanced than the narrative that has dominated in the past. Pretrained neural networks can embody an entirely different set of inductive biases from neural networks trained from scratch. Pretraining (especially if it involves meta-learning) can endow models with biases for few-shot inference, compositionality, and other powerful rule-like forms of behavior that until now have been assumed to require explicit symbolic mechanisms.  

\section{A new research agenda}
If we are beginning to answer classic questions about the symbolic capacity of neural networks, where should the field go next? Rather than seeing the progress of neural networks in capturing compositionality, productivity, and inductive biases as the end of such inquiries, 
we see the need for a new, more cognitively-oriented research agenda  focused on understanding the details of human behavior. 
This new agenda entails four directions.

\textbf{More informative diagnostic tasks.} Empirical progress will require us to develop more strongly-diagnostic tasks linked to  symbolic representations. 
Such tasks should be sufficiently different from tasks in LLM training data, and they should provide ``species-fair'' comparisons that put humans and machines on relatively even footing \cite{firestone2020performance}. Some recent tasks fall within this scope, including learning rule-like transformations on visual grids \cite{Chollet2019, HARC2024}, learning pseudolanguages for composing input-output rules \cite{LakeMLC}, and learning rules for list manipulation \cite{RuleMetaPrograms}
or for categorizing objects \cite{Piantadosi2016c}. A full comparison between humans and neural networks should consider accuracy, error patterns, implicit knowledge (reaction time, uncertainty, etc.), and explicit (verbalizable) knowledge, all of which a complete cognitive model should ideally capture. Many more tasks for comparing human and machine symbolic behavior will be needed too, given these networks' increasingly wide behavioral scope. 

\textbf{Mechanistic understanding.} Although stronger diagnostics will help identify  the most successful models, these models will not provide theoretical value unless they can be interpreted. This will require applying new interpretability
tools developed in AI research to explore how models 
internally represent symbolic computations (Box 3).  Interpretability research often starts with methods that are familiar to cognitive scientists and neuroscientists, such as examining the internal representations of models, but also uses sophisticated tools that are impractical to apply to humans, such as causal interventions on representations. Applying these tools to models that demonstrate symbolic capacities provides an opportunity to 
determine how symbolic information is represented and processed within those networks. This might take a variety of forms \cite{smolensky1988proper,mcgrath2024can}; neural networks might simply implement traditional symbolic systems \cite{Fodor1988,Marcus1998}, or they might be so qualitatively different from symbolic systems that they necessitate eliminating symbols from cognitive theories \cite{Rumelhart1986past,McClelland2010}.
The reality likely falls somewhere in between \cite{smolensky1988proper,smolensky2006harmonic,mcgrath2024can}, and
determining where today's neural networks fall on this spectrum will have important implications for the conclusions cognitive scientists draw from these systems.  
This effort requires not only technical advances in understanding neural networks but also theoretical and philosophical understanding of when we should conclude that one type of system is implementing another.

\textbf{Developmental models trained on human-like inputs.}
Tracing the origins of symbol-like representations in the mind, and how they support symbolic behavior, requires a developmental account. The massive amounts of training data they require make current  LLMs  developmentally implausible 
\cite{frank2023bridging}. To  address this, neural networks have been trained on child-scale inputs \cite{BabyLMReport,Huebner2021babyberta} or on inputs from egocentric video recordings \cite{,WangSAYCam2023,Zhuang2021a,Orhan2023NMI,Vong2024Science}, but these models  do not yet have strong symbolic capabilities. Contemporary LLMs also make extensive use of symbolic systems during training, either for pre-training (e.g., computer code), metalearning \cite{LakeMLC,mccoy2025modeling}, or post-training (e.g., symbolic verifiers for reasoning \cite{guo2025deepseek}). Children get experience with symbolic systems too, especially in formal education, but this type of experience still differs fundamentally from LLM training (e.g., pre-training on code is not a realistic proxy). Is extensive experience with symbolic systems needed for neural networks to develop compositionality, productivity, and the right kinds of inductive biases? Is there a different route to these capabilities that is more developmentally grounded?

\textbf{Cognitive models for predicting and explaining human behavior.}
Finally, the sub-goals above coalesce in a larger effort to build cognitive models using modern neural network tools. Beyond capturing compositionality, productivity and inductive biases at a coarse scale, a successful cognitive model should provide fine-grained predictions about human behavior. Such a model's internal mechanisms should be taken seriously as hypotheses about human mental processes. Getting there will require the ingredients mentioned above: advances in diagnostic tasks, mechanistic understanding, and developmentally plausible training.
If today's architectures are insufficient, architectural innovations could provide more data-efficient (and less symbolic-system-informed) learning~\cite{Webb2024}. Or innovations in training could help models better match the nuances of natural learning \cite{Smith2022,davidson2025goals,WangMinnow}. Finally, new classes of cognitive models could be developed by fine-tuning LLMs on human behavioral data from psychological studies \cite{BinzCentaur}. While this approach does not necessarily guarantee cognitive insights, it surfaces an important theoretical question which bears on the entirety of this new research agenda -- to what extent does behavioral alignment necessitate mechanistic alignment?

\section{Concluding remarks}

So, whither symbols? We have argued that the impressive capabilities of modern neural networks pose a challenge for symbolic systems as explanations of human cognition at the algorithmic level. Namely, the behaviors that have been used to support symbolic explanations in the past are now produced not just by humans, but by subsymbolic neural networks. In making this argument, we are not claiming that AI systems based on neural networks match human cognition in coherence, veracity, or inductive biases -- there are clear gaps for each, as well as ways in which the behavior of these AI systems is simply different from that of people (e.g., \cite{mccoy2024embers}). We are also not claiming that symbolic systems are irrelevant for understanding cognition -- one very reasonable resolution to this long-standing debate is to recognize that symbolic systems play an important explanatory role at the computational level, a notion that is reinforced by the fact that modern neural networks only produce phenomena reminiscent of human cognition when trained on data generated by symbolic systems. Our key claim is that support for symbolic systems at the algorithmic level is going to require new sources of evidence that go beyond those that have been used historically, or a deeper understanding of how neural networks produce these behaviors by implementing symbolic systems. Exploring these possibilities offers exciting new opportunities for cognitive science. 

\section{Acknowledgments}
We thank Solim LeGris for comments on an earlier version of this manuscript. EP was supported by a Young Faculty Award from the Defense Advanced
Research Projects Agency Grant \#D24AP00261, and is a paid consultant for Google Deepmind. The content of this article does not necessarily reflect that of the US Government or of Google and no official endorsement of this work should be inferred.

\vfill

\pagebreak

\setcounter{section}{0}


\section{Box 1: Levels of analysis} 

\label{box_levels}

Natural phenomena can be explained in different ways. To take an example used by David Marr, imagine we seek to understand bird flight. One kind of explanation might ask {\em why} birds can fly and focus on the general principles of aerodynamics and how birds' wings allow them to generate lift. Another kind of explanation might focus on {\em how} birds fly and focus on  feathers, muscles, and bones and how they work together to support flight. The same distinction applies when we seek to understand cognitive processes: one kind of explanation tries to identify the abstract principles underlying intelligent behavior, the other the mechanisms that actually make it possible. 

Marr expressed this distinction in terms of three levels of analysis at which we can ask questions in cognitive science~\cite{Marr1982}. At the {\em computational} level we ask what the abstract problem is that a system solves and what the ideal solution looks like. At the {\em algorithmic} level we ask how that solution might be approximated through a set of cognitive processes and what representations might be used in those processes. At the {\em implementation} level we ask how those representations and algorithms might be realized physically. 

Using a specific formalism does not necessarily make a commitment to a particular level of analysis. For example, formal logic might be used to characterize the solution to an abstract problem that human minds face, or a specific system of inference rules could be taken as a hypothesis about how people reason. Likewise, Bayes' rule can be used to specify the optimal solution to an inductive problem (assuming a particular set of inductive biases), or the process of evaluating hypotheses in terms of their plausibility and fit to data could be asserted as an algorithm people might follow for inductive inference. 

This ambiguity creates problems for interpreting claims about the role of symbolic systems in human cognition. When we see behavior consistent with logical reasoning or Bayesian inference, it is tempting to assume that this is a result of processes resembling the application of inference rules or Bayes' rule to symbolic hypotheses inside people's heads. However, consistency with a symbolic system could also be -- as we argue here -- a consequence of that symbolic system providing the correct computational-level account of the problem people are solving, and being extremely well-approximated (but not perfectly reproduced) by mechanisms that are more akin to artificial neural networks. 

In the main text we argue that many of the pieces of evidence for algorithmic-level explanations of human cognition in terms of symbolic systems are undermined by modern neural networks exhibiting similar behavior. But even if it is the case that subsymbolic explanations are correct at the algorithmic level, symbolic explanations can still be useful at the computational level. If we want to understand {\em why} a neural network -- natural or artificial -- behaves in a particular way, the fact that it is approximating a particular symbolic system provides valuable insight. Indeed, the neural networks that we discuss in the main text are all trained to approximate specific symbolic systems (including natural language, but also abstract formal languages, computer code, and math). This training seems critical to producing the behaviors we discuss, and approximating these symbolic systems is thus an important part of the way in which we explain these behaviors at the computational level. 

\vfill\pagebreak

\section{Box 2: Metalearning} \label{box_meta}
Meta-learning can expand the capabilities of traditional neural networks. Arguments favoring symbolic models point to generalizations (related to compositionality, productivity, strong inductive biases, etc.) that challenge traditional neural networks but are readily achievable with symbolic models. Recently, however, meta-learning has allowed neural networks to \emph{learn how to make more sophisticated generalizations} without altering today's architectures or their underlying principles such as distributed representations and gradient-based learning.

To contrast standard learning and meta-learning, standard (supervised) learning is about fitting a function $f(\cdot)$ (i.e., a neural network) that maps inputs $x_i$ to outputs $y_i$. The function is fit to the training set
$\mathcal{D}_{\text{train}} = \left\{ (x_i, y_i) \right\}_{i=1}^{n_{\text{train}}}$
with the hope of generalizing well to a test set,
$\mathcal{D}_{\text{test}} = \left\{ (x_j, y_j) \right\}_{j=1}^{n_{\text{test}}}$. 
However, generalization performance can be poor if the training set is too small (due to the network's generic inductive biases \cite{Geman1992}), or the test set differs from the training in systematic ways (e.g., novel compositions or longer outputs \cite{LakeBaroni2018generalization,Hupkes2020}).

\begin{figure}[b!]
    \centering
    \includegraphics[width=0.5\linewidth]{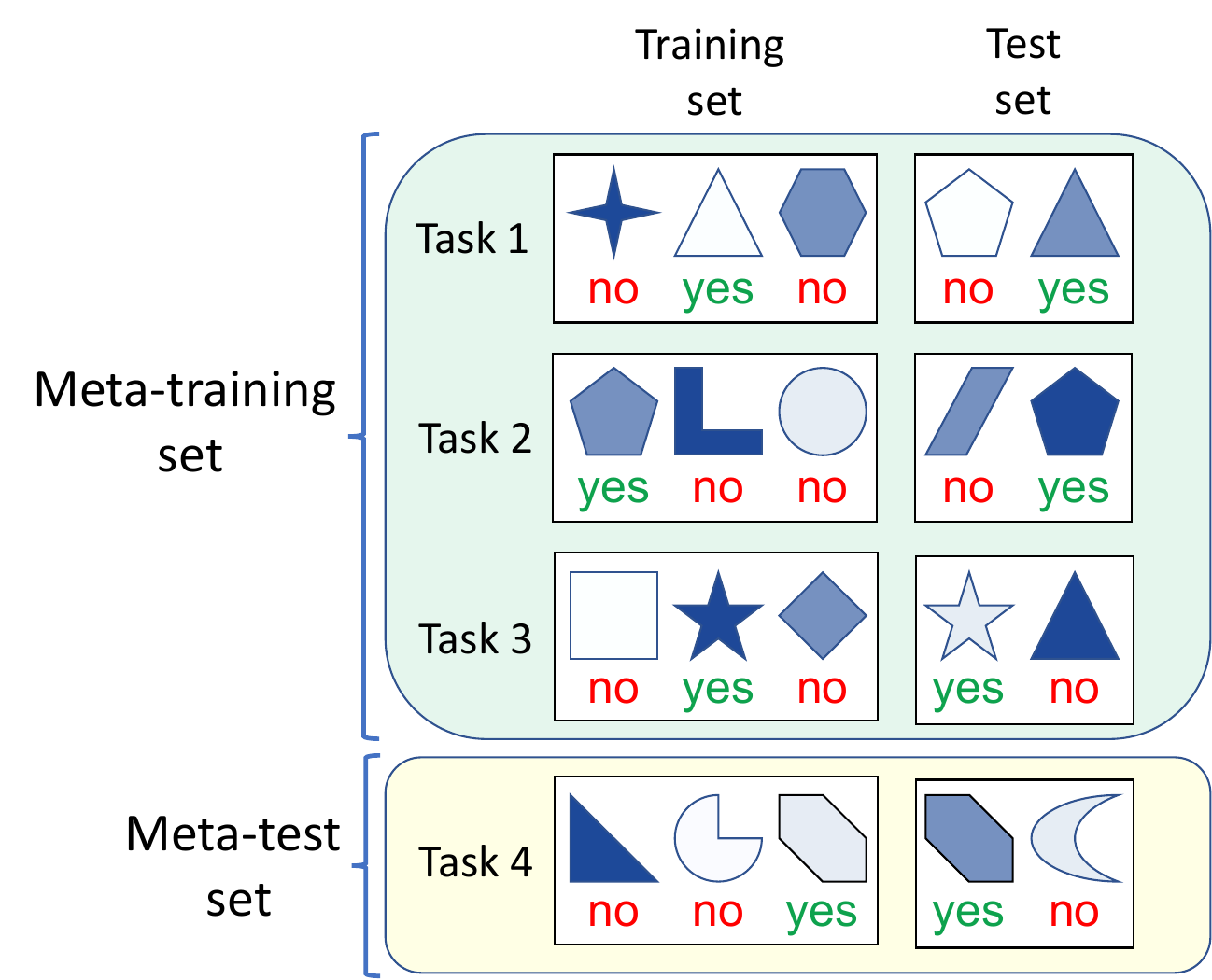}
    \caption{Meta-learning. In meta-learning, a system is shown a set of tasks called the meta-training set, and it is then tested on an additional set of tasks called the meta-test set. In the cases shown here, the system is performing visual category learning instantiated as binary classification, where the system must label whether each image belongs to the category (``yes'') or does not belong to the category (``no''). Each task $k$ comes with a training set $\mathcal{D}_{\text{train}}^{(k)}$ that illustrates the category, from which the system must learn a function $f(\cdot)^{(k)}$ that it then uses to label the items in the test set $\mathcal{D}_{\text{test}}^{(k)}$. Each task is one visual category to be learned. If a system meta-learns from the data shown here, it should end up with a shape bias---an inductive bias for categories defined by shape. In contrast, if the test set labels were all flipped, then the system should instead end up with a color bias (an inductive bias for categories defined by color). This illustration is inspired by \cite{Landau1988}.
}
    \label{fig:metalearning}
\end{figure}

Meta-learning offers a potential solution (see Figure \ref{fig:metalearning}). When standard learning leads to poor generalization, meta-learning can help by providing a series of standard learning problems so that the network can practice making better (compositional, productive, few-shot, etc.) generalizations from a given training set to a given test set. That is, meta-learning is about learning how to fit functions $f(\cdot)^{(k)}$ such that they generalize well from training set $\mathcal{D}_{\text{train}}^{(k)}$ to test set $\mathcal{D}_{\text{test}}^{(k)}$ for a set of distinct yet related learning problem $k= 1,\dots,K$ (called episodes). After meta-learning is complete, the system is evaluated on a new task $K+1$ and given $\mathcal{D}_{\text{train}}^{(K+1)}$ to induce $f(\cdot)^{(K+1)}$, should be better equipped to generalize to $\mathcal{D}_{\text{test}}^{(K+1)}$.

Meta-learning can be applied to neural networks in two main ways. If each task is allowed a different set of weights $\theta_k$, one kind of meta-learning (e.g., \cite{Finn2017maml}) aims to learn a powerful starting set of weights $\theta$ such that a small number of gradient steps based on  $\mathcal{D}_{\text{train}}^{(k)}$ induces a function $f(\cdot)^{(k)} = f_{\theta_k}(\cdot)$. Another kind of metalearning assumes a single neural network is used to perform all of the tasks, but changes its behavior based on the context (e.g., \cite{Wang2016b}). More formally, the neural network can induce a new function $f(\cdot)^{(k)}$ via a prompting-like mechanism by passing $\mathcal{D}_{\text{train}}^{(k)}$ as in-context input, i.e., $f(\cdot)^{(k)} = f_\theta(\cdot,\mathcal{D}_{\text{train}}^{(k)})$. In either form, a standard neural network that struggles to make the right kinds of generalizations can be enhanced through meta-learning which effectively provides it with the right incentives to improve those generalizations, and the right opportunities to practice making those generalizations \cite{IriePIP}.

For cognitive scientists, these two forms of metalearning provide a way to create systems that abstract shared structure across tasks. In the first form of metalearning the shared structure is instantiated in a set of initial weights, which act like a Bayesian prior distribution and support rapid learning by further modifying the weights through gradient descent \cite{Grant2019recasting,mccoy2025modeling}.  In the second form of metalearning the shared structure is instantiated in a set of weights that allow the network to adapt its behavior without additional changes to the weights, producing the phenomena of in-context learning discussed in the main text. This also has a Bayesian interpretation, with the weights representing a probability distribution over tasks that is conditioned based on the context information \cite{Muller2022transformers}. Both approaches thus modify the inductive biases of learning systems, but for two different kinds of learning.

\vfill\pagebreak

\section{Box 3: Mechanistic interpretability} 

\label{box_mechanistic}

Although large neural networks, in particular LLMs, are frequently referred to as ``black boxes'', there has been significant progress in understanding the internals of these systems. Such work was spearheaded by researchers in linguistics and cognitive science seeking to understand the structure of the latent representations of LLMs -- focusing, for example on syntactic categories and hierarchical structure \cite{belinkov2019analysis,linzen2021syntactic,pavlick2022semantic}. More recently, work on LLM interpretability has focused on \textit{mechanistic} explanations, which outline the specific (transformer) components responsible for implementing solutions to specific tasks.

Much of this work has focused on characterizing the role of individual \textit{attention heads} -- architectural components responsible for routing information across layers in the Transformer. For example, one important and reproducible finding is the emergence of induction heads, which specialize in the task of (largely content-independent) copying~\cite{olsson2022context} from the input into the output.

Components such as induction heads do not exist in isolation, but rather, appear to work in tandem with similarly specialized components within ``circuits'' which execute complex tasks.  Wang et al.\ \cite{wang2022interpretability} demonstrate how induction heads work together with specialized components for duplicate-detection and inhibition to correctly predict the indirect object in contexts such as ``When John and Mary went to the store, John gave a drink to ...''. Circuits of this type have been found for a range of key cognitive behaviors, including abstract reasoning \cite{yang2025emergent}, compositional generalization \cite{tang2025explainable}, and numeric comparisons \cite{grant2025emergent}, and even abstract visual reasoning \cite{lepori2024beyond}. These circuits are often brittle to human-perceived input variation (e.g., the indirect object circuit breaks when presented with small syntactic variations) but nonetheless the components themselves are often highly modular within the transformer. Circuits have been shown to reproduce in different tasks \cite{merullo2023circuit} and languages \cite{zhang}, and can be connected to more formal theories of transformer computation \cite{lindner2023tracr,weiss2021thinking}. 

Parallel work has focused on the structure and geometry of the high-dimensional embedding space in which LLMs operate. There is increasing evidence that the high-dimensional and non-linear embedding space of LLMs converges to make use of simple, linear, and sparse representations  of otherwise complex tasks and behaviors. For example, so-called ``function vectors'' \cite{Todd2023,merullo2024language,hendel2023context} can be extracted from the activation patterns associated with a handful of in-context learning instances, and subsequently injected into new contexts in order to influence model behavior--for example, getting a model to produce the antonym for each input word, even in the absence of explicit contextual cues to do so. Indeed, these representations can be directly interpreted as implementing forms of symbol processing~\cite{yang2025emergent} and symbolic representations \citep{mccoy2018rnns,mccoy2022implicit}. Similarly, multiple lines of recent work have emphasized the sparsity of the representation space learned by LLMs, highlighting ways that single neurons and low-dimensional representations can mediate behavior \cite{vig2020investigating,bau2020understanding,yu2023characterizing,meng2022locating}. For example, the recent interest in sparse autoencoders for revealing and controlling models' feature space \cite{lan2024sparse} suggests a quality to the representations that is, if not discrete, at least easier to reconcile with traditionally symbolic accounts of these behaviors.

\vfill\pagebreak

\bibliographystyle{unsrt}

\end{document}